\documentclass{article} 
\usepackage{nips12submit_e,times}
\usepackage[latin1]{inputenc}
\usepackage{amsmath}
\usepackage{amsfonts}
\usepackage{amssymb}
\usepackage{caption} 
\usepackage{cite} 
\usepackage{subcaption} 
\usepackage[pdftex]{graphicx}
\nipsfinalcopy

\title{Saturating Auto-Encoders} 

\author{
Rostislav Goroshin \thanks{The authors thank Joan Bruna and David Eigen for their useful suggestions and comments.} \\
Courant Institute of Mathematical Science\\
New York University \\
\texttt{goroshin@cs.nyu.edu} \\
\And
Yann LeCun \\
Courant Institute of Mathematical Science\\
New York University \\
\texttt{yann@cs.nyu.edu} \\
\date{} }

\begin{document}
\maketitle

\begin{abstract} 
We introduce a simple new regularizer for auto-encoders whose hidden-unit activation functions contain at least one zero-gradient (saturated) region. This regularizer explicitly encourages activations in the saturated region(s) of the corresponding activation function. We call these Saturating Auto-Encoders (SATAE). We show that the saturation regularizer explicitly limits the SATAE's ability to reconstruct inputs which are not near the data manifold. Furthermore, we show that a wide variety of features can be learned when different activation functions are used. Finally, connections are established with the Contractive and Sparse Auto-Encoders.    

\end{abstract} 
\section{Introduction} 
An auto-encoder is a conceptually simple neural network used for obtaining useful data representations through unsupervised training. It is composed of an encoder which outputs a hidden (or latent) representation and a decoder which attempts to reconstruct the input using the hidden representation as its input. Training consists of minimizing a reconstruction cost such as $L_2$ error. However this cost is merely a proxy for the true objective: to obtain a useful latent representation. Auto-encoders can  implement many dimensionality reduction techniques such as PCA and Sparse Coding (SC) \cite{DHS}\cite{SC}\cite{LISTA}. This makes the study of auto-encoders very appealing from a theoretical standpoint. In recent years, renewed interest in auto-encoders networks has mainly been due to their empirical success in unsupervised feature learning \cite{SAE1}\cite{SAE2}\cite{CAE}\cite{DAE}. \\

\noindent
When minimizing only reconstruction cost, the standard auto-encoder does not typically learn any meaningful hidden representation of the data. Well known theoretical and experimental results show that a linear auto-encoder with trainable encoding and decoding matrices, $W^e$ and $W^d$ respectively, learns the identity function if $W^e$ and $W^d$ are full rank or over-complete. The linear auto-encoder learns the principle variance directions (PCA) if $W^e$ and $W^d$ are rank deficient \cite{DHS}. It has been observed that other representations can be obtained by regularizing the latent representation. This approach is exemplified by the Contractive and Sparse Auto-Encoders \cite{CAE} \cite{SAE1} \cite{SAE2}. Intuitively, an auto-encoder with limited capacity will focus its resources on reconstructing portions of the input space in which data samples occur most frequently. From an energy based perspective, auto-encoders achieve low reconstruction cost in portions of the input space with high data density (recently, \cite{bengio_new} has examined this perspective in depth). If the data occupies some low dimensional manifold in the higher dimensional input space then minimizing reconstruction error achieves low energy on this manifold. Useful latent state regularizers raise the energy of points that do not lie on the manifold, thus playing an analogous role to minimizing the partition function in maximum likelihood models. In this work we introduce a new type of regularizer that does this explicitly for auto-encoders with a non-linearity that contains at least one flat (zero gradient) region. We show examples where this regularizer and the choice of nonlinearity determine the feature set that is learned by the auto-encoder.      

\section{Latent State Regularization}  
Several auto-encoder variants which regularize their latent states have been proposed, they include the sparse auto-encoder and the contractive auto-encoder\cite{SAE1}\cite{SAE2}\cite{CAE}. The sparse auto-encoder includes an over-complete basis in the encoder and imposes a sparsity inducing (usually $L_1$) penalty on the hidden activations. This penalty prevents the auto-encoder from learning to reconstruct all possible points in the input space and focuses the expressive power of the auto-encoder on representing the data-manifold. Similarly, the contractive auto-encoder avoids trivial solutions by introducing an auxiliary penalty which measures the square  Frobenius norm of the Jacobian of the latent representation with respect to the inputs. This encourages a constant latent representation except around training samples where it is counteracted by the reconstruction term. It has been noted in \cite{CAE} that these two approaches are strongly related. The contractive auto-encoder explicitly encourages small entries in the Jacobian, whereas the sparse auto-encoder is encouraged to produce mostly zero (sparse) activations which can be designed to correspond to mostly flat regions of the nonlinearity, thus also yielding small entries in the Jacobian.

\subsection{Saturating Auto-Encoder through Complementary Nonlinearities}     
\begin{figure}
\centering 
\includegraphics[scale=0.6]{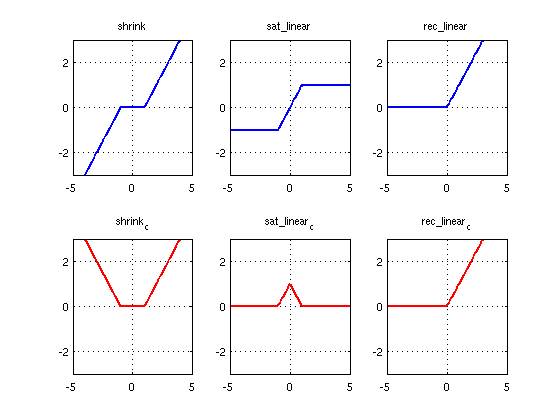}
\caption{Three nonlinearities (top) with their associated complementary regularization functions(bottom).}  
\label{fig:nonlin}
\end{figure}

Our goal is to introduce a simple new regularizer which explicitly raises reconstruction error for inputs not near the data manifold. Consider activation functions with at least one flat region; these include shrink, rectified linear, and saturated linear (Figure~\ref{fig:nonlin}). Auto-encoders with such nonlinearities lose their ability to accurately reconstruct inputs which produce activations in the zero-gradient regions of their activation functions. Let us denote the auto-encoding function $x_r = G(x,W)$, $x$ being the input, $W$ the trainable parameters in the auto-encoder, and $x_r$ the reconstruction. One can define an energy surface
through the reconstruction error:
\[
  E_W(x) = ||x-G(x,W)||^2
\]
Let's imagine that $G$ has been trained to produce a low reconstruction error at a particular data point $x^*$. If $G$ is constant when $x$ varies along a particular direction $v$, then the energy will grow quadratically along that particular direction as $x$ moves away from $x^*$. If $G$ is trained to produce low reconstruction errors on a set of samples while being subject to a regularizer that tries to make it constant in as many directions as possible, then the reconstruction energy will act as a {\em contrast function} that will take low values around areas of high data density and larger values everywhere else (similarly to a negative log likelihood function for a density estimator).

The proposed auto-encoder is a simple implementation of this idea. 
Using the notation $W =\{W^e,B^e,W^d,B^d\}$, the auto-encoder function is defined as
\[
  G(x,W) = W^d F(W^e x+B^e) + B^d
\]
where $W^e$, $B^e$, $W^d$, and $B^d$ are the encoding matrix, encoding bias, decoding matrix, and decoding bias, respectively, and $F$ is the vector function that applies the scalar function $f$ to each of its components. $f$ will be designed to have "flat spots", i.e. regions where the derivative is zero (also referred to as the saturation region).

The loss function minimized by training is the sum of the reconstruction energy $E_W(x)=||x-G(x,W)||^2$ and a term that pushes the components of $W^e x + B^e$ towards the flat spots of $f$. This is performed through the use of a {\em complementary function} $f_c$, associated with the non-linearity $f(z)$. The basic idea is to design $f_c(z)$ so that its value corresponds to the distance of $z$ to one of the flat spots of $f(z)$. Minimizing $f_c(z)$ will push $z$ towards the flat spots of $f(z)$. With this in mind, we introduce a penalty of the form $f_c(\sum_{j=1}^d W^e_{ij}x_j + b^e_i)$ which encourages the argument to be in the saturation regime of the activation function ($f$). We refer to auto-encoders which include this regularizer as Saturating Auto-Encoders (SATAEs). For activation functions with zero-gradient regime(s) the complementary nonlinearity ($f_c$) can be defined as the distance to the nearest saturation region. Specifically, let $S = \{z \mid  f'(z) = 0\}$ then we define $f_c(z)$ as: 

\begin{equation}
f_c(z) = \inf_ {z' \in S} |z-z'|.   
\end{equation}   

\noindent
Figure 1 shows three activation functions and their associated complementary nonlinearities. The complete loss to be minimized by a SATAE with nonlinearity $f$ is: 

\begin{equation} 
L = \sum_{x \in D} \frac{1}{2} \|x-\left(W^d F(W^e x+B^e)+B^d\right)\|^2 + \alpha \sum_{i=1}^{d_h}f_c(W^e_i x + b^e_i),
\end{equation}    

\noindent
where $d_h$ denotes the number of hidden units. The hyper-parameter $\alpha$ regulates the trade-off between reconstruction and saturation.  

\section{Effect of the Saturation Regularizer} 
We will examine the effect of the saturation regularizer on auto-encoders with a variety of activation functions. It will be shown that the choice of activation function is a significant factor in determining the type of basis the SATAE learns. First, we will present results on toy data in two dimensions followed by results on higher dimensional image data.

\subsection{Visualizing the Energy Landscape}  
\begin{figure}
\centering 
\includegraphics[scale=0.25]{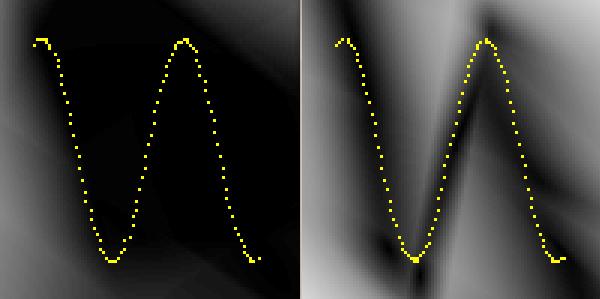}
\caption{Energy surfaces for unregularized (left), and regularized (right) solutions obtained on SATAE-shrink and 10 basis vectors. Black corresponds to low reconstruction energy. Training points lie on a one-dimensional manifold shown in yellow.}  
\label{fig:toyshrink}
\end{figure}

\begin{figure}
\centering 
\includegraphics[scale=0.25]{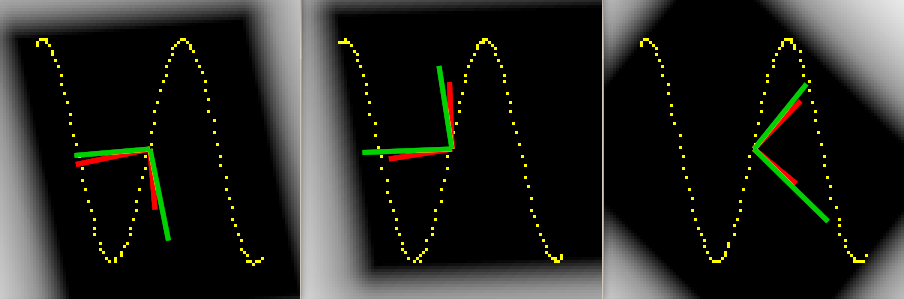}
\includegraphics[scale=0.25]{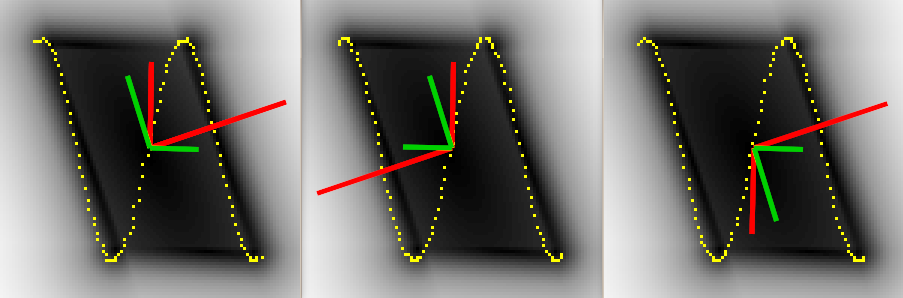}
\caption{SATAE-SL toy example with two basis elements. Top Row: three randomly initialized solutions obtained with no regularization. Bottom Row: three randomly initialized solutions obtained with regularization.}  
\label{fig:toysatlinear} 
\end{figure}

Given a trained auto-encoder the reconstruction error can be evaluated for a given input $x$. For low-dimensional spaces ($\mathbb{R}^n$, where $n \leq 3$) we can evaluate the reconstruction error on a regular grid in order to visualize the portions of the space which are well represented by the auto-encoder. More specifically we can compute $E(x) = \frac{1}{2} \|x - x_r \|^2$ for all $x$ within some bounded region of the input space. Ideally, the reconstruction energy will be low for all $x$ which are in the training set and high elsewhere. Figures~\ref{fig:toyshrink} and~\ref{fig:toysatlinear} depict the resulting reconstruction energy for inputs $x \in \mathbb{R}^2$, and  $-1 \leq x_i \leq 1$. Black corresponds to low reconstruction energy. The training data consists of a one dimensional manifold shown overlain in yellow. Figure~\ref{fig:toyshrink} shows a toy example for a SATAE which uses ten basis vectors and a shrink activation function. Note that adding the saturation regularizer decreases the volume of the space which is well reconstructed, however good reconstruction is maintained on or near the training data manifold. The auto-encoder in Figure~\ref{fig:toysatlinear} contains two encoding basis vectors (red), two decoding basis vectors (green), and uses a saturated-linear activation function. The encoding and decoding bases are unconstrained. The unregularized auto-encoder learns an orthogonal basis with a random orientation. The region of the space which is well reconstructed corresponds to the outer product of the linear regions of two activation functions; beyond that the error increases quadratically with the distance. Including the saturation regularizer induces the auto-encoder basis to align with the data and to operate in the saturation regime at the extreme points of the training data, which limits the space which is well reconstructed. Note that because the encoding and decoding weights are separate and unrestricted, the encoding weights were scaled up to effectively reduce the width of the linear regime of the nonlinearity. 

\subsection{SATAE-shrink}
Consider a SATAE with a shrink activation function and shrink parameter $\lambda$. The corresponding complementary nonlinearity, derived using Equation 1 is given by: 
\begin{equation} 
\nonumber
shrink_c(x) =
\begin{cases}
abs(x), \text{ } |x| > \lambda\\
0, \text{ elsewhere}
\end{cases}.
\end{equation} 

\begin{figure}
\centering 
\begin{subfigure}[b]{0.225\textwidth}
		\centering 
		\includegraphics[width=2.5cm, height=10cm]{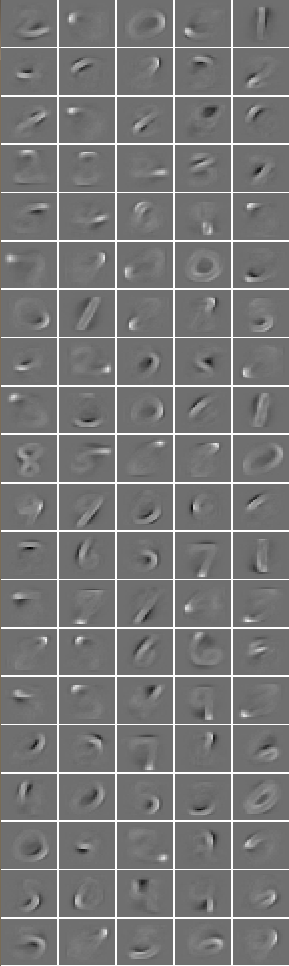}
		\caption{}
	\end{subfigure} 
	\begin{subfigure}[b]{0.225\textwidth}
		\centering 
		\includegraphics[width=2.5cm, height=10cm]{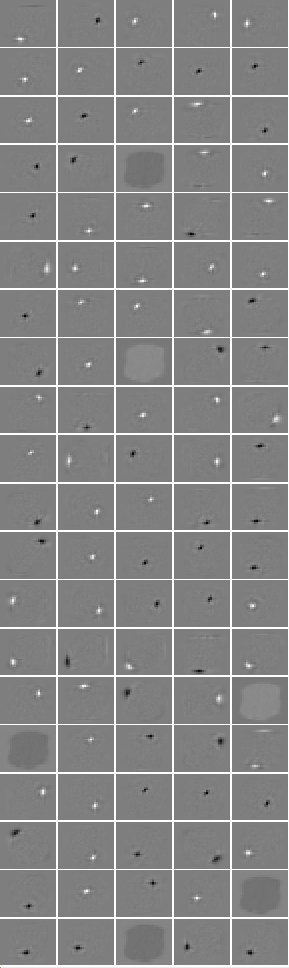}
		\caption{}
	\end{subfigure} 
	\begin{subfigure}[b]{0.2\textwidth}
		\centering 
		\includegraphics[width=2.5cm, height=10cm]{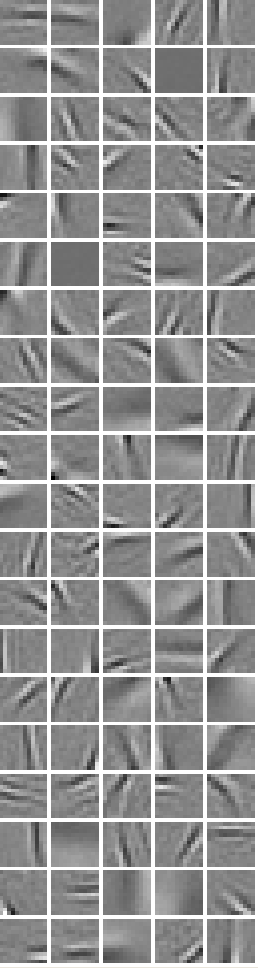}
		\caption{}
	\end{subfigure} 
	\begin{subfigure}[b]{0.2\textwidth}
		\centering 
		\includegraphics[width=2.5cm, height=10cm]{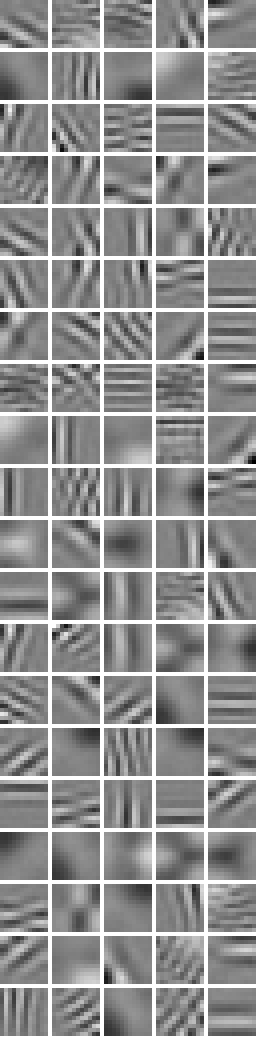}
		\caption{}
	\end{subfigure} \\
	\begin{subfigure}[b]{0.225\textwidth}
		\centering 
		\includegraphics[width=2.5cm, height=10cm]{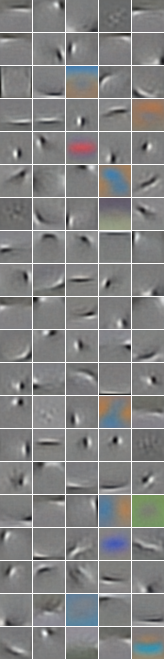}
		\caption{}
	\end{subfigure} 
	\begin{subfigure}[b]{0.225\textwidth}
		\centering 
		\includegraphics[width=2.5cm, height=10cm]{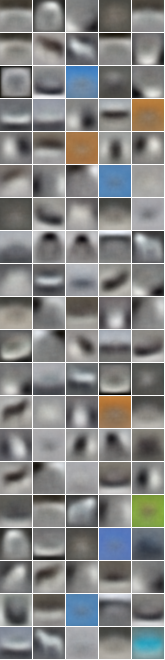}
		\caption{}
	\end{subfigure} 
	\begin{subfigure}[b]{0.2\textwidth}
		\centering 
		\includegraphics[width=2.5cm, height=10cm]{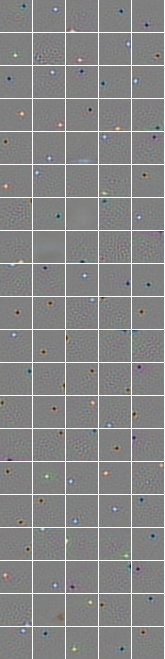}
		\caption{}
	\end{subfigure} 
	\begin{subfigure}[b]{0.2\textwidth}
		\centering 
		\includegraphics[width=2.5cm, height=10cm]{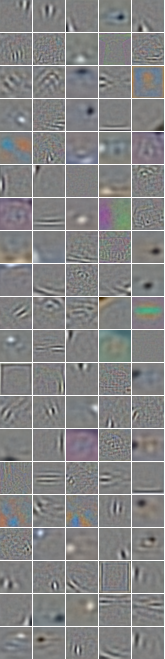}
		\caption{}
	\end{subfigure} 
\caption{Basis elements learned by the SATAE using different nonlinearities on: 28x28 binary MNIST digits, 12x12 gray scale natural image patches, and CIFAR-10. (a) SATAE-shrink trained on MNIST, (b) SATAE-saturated-linear trained on MNIST, (c) SATAE-shrink trained on natural image patches, (d) SATAE-saturated-linear trained on natural image patches, (e)-(f) SATAE-shrink trained on CIFAR-10 with $\alpha=0.1$ and $\alpha=0.5$, respectively, (g)-(h) SATAE-SL trained on CIFAR-10 with $\alpha=0.1$ and $\alpha=0.6$, respectively.  }
\label{fig:results}
\end{figure}

Note that $shrink_c(W^e x + b^e) =  abs(shrink(W^e x + b^e))$, which corresponds to an $L_1$ penalty on the activations. Thus this SATAE is equivalent to a sparse auto-encoder with a shrink activation function. Given the equivalence to the sparse auto-encoder we anticipate the same scale ambiguity which occurs with $L_1$ regularization. This ambiguity can be avoided by normalizing the decoder weights to unit norm. It is expected that the SATAE-shrink will learn similar features to those obtained with a sparse auto-encoder, and indeed this is what we observe. Figure~\ref{fig:results}(c) shows the decoder filters learned by an auto-encoder with shrink nonlinearity trained on gray-scale natural image patches. One can recognize the expected Gabor-like features when the saturation penalty is activated. When trained on the binary MNIST dataset the learned basis is comprised of portions of digits and strokes. Nearly identical results are obtained with a SATAE which uses a rectified-linear activation function. This is because a rectified-linear function with an encoding bias behaves as a positive only shrink function, similarly the complementary function is equivalent to a positive only $L_1$ penalty on the activations.

\begin{figure}
\centering 
\begin{subfigure}[b]{0.15\textwidth}
		\centering 
		\includegraphics[scale=2]{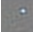}\\
		\includegraphics[scale=2]{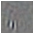} 
	\end{subfigure} 
	\begin{subfigure}[b]{0.15\textwidth}
		\centering 
		\includegraphics[scale=2]{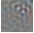} \\
		\includegraphics[scale=2]{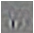} 
	\end{subfigure} 
	\begin{subfigure}[b]{0.15\textwidth}
		\centering 
		\includegraphics[scale=2]{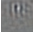}\\
		\includegraphics[scale=2]{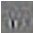} 
		
	\end{subfigure} 
	\begin{subfigure}[b]{0.15\textwidth}
		\centering 
		\includegraphics[scale=2]{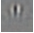} \\
		\includegraphics[scale=2]{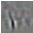} 
	\end{subfigure} 
	\begin{subfigure}[b]{0.15\textwidth}
		\centering 
		\includegraphics[scale=2]{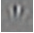}
		\includegraphics[scale=2]{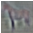} \\
		
	\end{subfigure} 
\caption{Evolution of two filters with increasing saturation regularization for a SATAE-SL trained on CIFAR-10. Filters corresponding to larger values of $\alpha$ were initialized using the filter corresponding to the previous $\alpha$. The regularization parameter was varied from 0.1 to 0.5 (left to right) in the top five images and 0.5 to 1 in the bottom five }
\label{fig:horse}
\end{figure} 

\subsection{SATAE-saturated-linear} 
Unlike the SATAE-shrink, which tries to compress the data by minimizing the number of active elements; the SATAE saturated-linear (SATAE-SL) tries to compress the data by encouraging the latent code to be as close to binary as possible. Without a saturation penalty this auto-encoder learns to encode small groups of neighboring pixels. More precisely, the auto-encoder learns the identity function on all datasets. An example of such a basis is shown in Figure \ref{fig:results}(b). With this basis the auto-encoder can perfectly reconstruct any input by producing small activations which stay within the linear region of the nonlinearity. Introducing the saturation penalty does not have any effect when training on binary MNIST. This is because the scaled identity basis is a global minimizer of Equation 2 for the SATAE-SL on any binary dataset. Such a basis can perfectly reconstruct any binary input while operating exclusively in the saturated regions of the activation function, thus incurring no saturation penalty. On the other hand, introducing the saturation penalty when training on natural image patches induces the SATAE-SL to learn a more varied basis (Figure \ref{fig:results}(d)). 

\subsection{Experiments on CIFAR-10}
SATAE auto-encoders with 100 and 300 basis elements were trained on the CIFAR-10 dataset, which contains small color images of objects from ten categories. In all of our experiments the auto-encoders were trained by progressively increasing the saturation penalty (details are provided in the next section). This allowed us to visually track the effect of the saturation penalty on individual basis elements. Figure \ref{fig:results}(e)-(f) shows the basis learned by SATAE-shrink with small and large saturation penalty, respectively. Increasing the saturation penalty has the expected effect of reducing the number of nonzero activations. As the saturation penalty increases, active basis elements become responsible for reconstructing a larger portion of the input. This induces the basis elements to become less spatially localized. This effect can be seen by comparing corresponding filters in Figure \ref{fig:results}(e) and (f). Figures \ref{fig:results}(g)-(h)  show the basis elements learned by SATAE-SL with small and large saturation penalty, respectively. The basis learned by SATAE-SL with a small saturation penalty resembles the identity basis, as expected (see previous subsection). Once the saturation penalty is increased small activations become more heavily penalized. To increase their activations the encoding basis elements may increase in magnitude or align themselves with the input. However, if the encoding and decoding weights are tied (or fixed in magnitude) then reconstruction error would increase if the weights were merely scaled up. Thus the basis elements are forced to align with the data in a way that also facilitates reconstruction. This effect is illustrated in Figure \ref{fig:horse} where filters corresponding to progressively larger values of the regularization parameter are shown. The top half of the figure shows how an element from the identity basis ($\alpha=0.1$) transforms to a localized edge ($\alpha=0.5$). The bottom half of the figure shows how a localized edge ($\alpha=0.5$) progressively transforms to a template of a horse ($\alpha=1$).

\section{Experimental Details}
Because the regularizer explicitly encourages activations in the zero gradient regime of the nonlinearity, many encoder basis elements would not be updated via back-propagation through the nonlinearity if the saturation penalty were large. In order to allow the basis elements to deviate from their initial random states we found it necessary to progressively increase the saturation penalty. In our experiments the weights obtained at a minimum of Equation 2 for a smaller value of $\alpha$ were used to initialize the optimization for a larger value of $\alpha$. Typically, the optimization began with $\alpha=0$ and was progressively increased to $\alpha=1$ in steps of $0.1$. The auto-encoder was trained for 30 epochs at each value of $\alpha$. This approach also allowed us to track the evolution of basis elements as a function of $\alpha$ (Figure \ref{fig:horse}). In all experiments data samples were normalized by subtracting the mean and dividing by the standard deviation of the dataset. The auto-encoders used to obtain the results shown in Figure \ref{fig:results} (a),(c)-(f) used 100 basis elements, others used 300 basis elements. Increasing the number of elements in the basis did not have a strong qualitative effect except to make the features represented by the basis more localized. The decoder basis elements of the SATAEs with shrink and rectified-linear nonlinearities were reprojected to the unit sphere after every 10 stochastic gradient updates. The SATAEs which used saturated-linear activation function were trained with tied weights.  All results presented were obtained using stochastic gradient descent with a constant learning rate of 0.05.

\section{Discussion}

In this work we have introduced a general and conceptually simple latent state regularizer. It was demonstrated that a variety of feature sets can be obtained using a single framework. The utility of these features depend on the application. In this section we extend the definition of the saturation regularizer to include functions without a zero-gradient region. The relationship of SATAEs with other regularized auto-encoders will be discussed. We conclude with a discussion on future work.   

\subsection{Extension to Differentiable Functions}
We would like to extend the saturation penalty definition (Equation 1) to differentiable functions without a zero-gradient region. An appealing first guess for the complimentary function is some positive function of the first derivative, $f_c(x) = |f'(x)|$ for instance. This may be an appropriate choice for monotonic activation functions which have their lowest gradient regions at the extrema (e.g. sigmoids). However some activation functions may contain regions of small or zero gradient which have negligible extent, at the extrema for instance. We would like our definition of the complimentary function to not only measure the local gradient in some region, but to also measure it's extent. For this purpose we employ the concept of average variation over a finite interval. We define the average variation of $f$ at $x$ in the positive and negative directions at scale $l$, respectively as: 

\begin{eqnarray}
\nonumber
\Delta_l^+ f(x) &=& \frac{1}{l} \int_x ^{x+l} |f'(u)| du = |f'(x)| * \Pi_l^+(x)\\
\nonumber
\Delta_l^- f(x) &=& \frac{1}{l} \int_{x-l} ^x |f'(u)| du = |f'(x)| * \Pi_l^-(x).
\end{eqnarray} 

Where $*$ denotes the continuous convolution operator. $\Pi_l^+(x)$ and $\Pi_l^-(x)$ are uniform averaging kernels in the positive and negative directions, respectively. Next, define a directional measure of variation of $f$ by integrating the average variation at all scales. 

\begin{eqnarray}
\nonumber
M^+ f(x) &=& \int_0^{+\infty} \Delta_l^+ f(x) w(l)dl = \left[\int_0^{+\infty} w(l) \Pi^+_l(x) dl \right] * |f'(x)| \\
\nonumber
M^- f(x) &=& \int_0^{+\infty} \Delta_l^- f(x) w(l)dl = \left[\int_0^{+\infty} w(l) \Pi^-_l(x) dl \right] * |f'(x)| .
\end{eqnarray} 

Where $w(l)$ is chosen to be a sufficiently fast decreasing function of $l$ to insure convergence of the integral. The integral with which $|f'(x)|$ is convolved in the above equation evaluates to some decreasing function of $x$ for $\Pi^+$ with support $x \geq 0$. Similarly, the integral involving $\Pi^-$ evaluates to some increasing function of $x$ with support $x \leq 0$. This function will depend on $w(l)$. The functions $M^+f(x)$ and $M^-f(x)$ measure the average variation of $f(x)$ at all scales $l$ in the positive and negative direction, respectively. We define the complimentary function $f_c(x)$ as: 

\begin{equation} 
f_c(x) = min(M^+f(x),M^-f(x)).
\end{equation} 

\begin{figure}
\centering 
\includegraphics[scale=0.43]{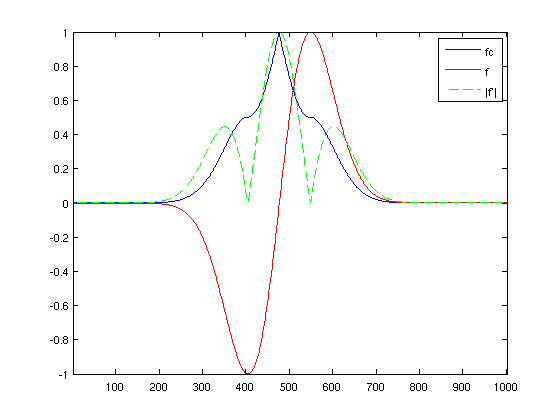}
\caption{Illustration of the complimentary function ($f_c$) as defined by Equation 3 for a non-monotonic activation function ($f$). The absolute derivative of $f$ is shown for comparison.}  
\label{fig:diff_cc}
\end{figure} 

An example of a complimentary function defined using the above formulation is shown in Figure~\ref{fig:diff_cc}. Whereas $|f'(x)|$ is minimized at the extrema of $f$, the complimentary function only plateaus at these locations.

\subsection{Relationship with the Contractive Auto-Encoder} 
Let $h_i$ be the output of the $i^{th}$ hidden unit of a single-layer auto-encoder with point-wise nonlinearity $f(\cdot)$. The regularizer imposed by the contractive auto-encoder (CAE) can be expressed as follows: 

\begin{equation}
\nonumber
\sum_{ij} \left(\frac{\partial h_i}{\partial x_j} \right)^2 = \sum_i ^{d_h} \left(f'(\sum_{j=1}^d W^e_{ij}x_j + b_i)^2 \| W^e_i \| ^2 \right),
\end{equation}  
 
\noindent
where $x$ is a $d$-dimensional data vector, $f'(\cdot)$ is the derivative of $f(\cdot)$, $b_i$ is the bias of the $i^{th}$ encoding unit, and $W^e_i$ denotes the $i^{th}$ row of the encoding weight matrix. The first term in the above equation tries to adjust the weights so as to push the activations into the low gradient (saturation) regime of the nonlinearity, but is only defined for differentiable activation functions. Therefore the CAE indirectly encourages operation in the saturation regime. Computing the Jacobian, however, can be cumbersome for deep networks. Furthermore, the complexity of computing the Jacobian is $O(d \times d_h)$, although a more efficient implementation is possible \cite{CAE}, compared to the $O(d_h)$ for the saturation penalty.

\subsection{Relationship with the Sparse Auto-Encoder}
In Section 3.2 it was shown that SATAEs with shrink or rectified-linear activation functions are equivalent to a sparse auto-encoder. Interestingly, the fact that the saturation penalty happens to correspond to $L_1$ regularization in the case of SATAE-shrink agrees with the findings in \cite{LISTA}. In their efforts to find an architecture to approximate inference in sparse coding, Gregor et al. found that the shrink function is particularly compatible with $L_1$ minimization. Equivalence to sparsity only for some activation functions suggests that SATAEs are a generalization of sparse auto-encoders. Like the sparsity penalty, the saturation penalty can be applied at any point in a deep network for the same computational cost. However, unlike the sparsity penalty the saturation penalty is adapted to the nonlinearity of the particular layer to which it is applied. 

\subsection{Future Work} 
We intend to experimentally demonstrate that the representations learned by SATAEs are useful as features for learning common tasks such as classification and denoising. We will also address several open questions, namely: (i) how to select (or learn) the width parameter ($\lambda$) of the nonlinearity, and (ii) how to methodically constrain the weights. We will also explore SATAEs that use a wider class of non-linearities and architectures.

\end{document}